\documentclass[lettersize,journal]{IEEEtran}
\usepackage{amsmath,amsfonts}
\usepackage{algorithmic}
\usepackage{algorithm}
\usepackage{array}
\usepackage[caption=false,font=normalsize,labelfont=sf,textfont=sf]{subfig}
\usepackage{textcomp}
\usepackage{stfloats}
\usepackage{url}
\usepackage{verbatim}
\usepackage{graphicx}
\usepackage{cite}
\usepackage{orcidlink}
\usepackage{xcolor}
\usepackage{multirow}
\usepackage{hyperref}
\usepackage{arydshln}

\hypersetup{hidelinks,
	colorlinks=true,
	pdfstartview=Fit,
	breaklinks=true}

\begin{document}

\title{\textit{UniVCD}: A New Method for Unsupervised Change Detection in the Open-Vocabulary Era}

\author{
	Ziqiang~Zhu,
	Bowei~Yang,

	\thanks{
	Z. Zhu and B. Yang are with the School of Aeronautics and Astronautics, Zhejiang University, Hangzhou, 310027, P.R.China (e-mail: \{ziqiangzhu, boweiy\}@zju.edu.cn). B. Yang is the corresponding author.
	}
}

\markboth
{Zhu \MakeLowercase{\textit{et al.}}: \textit{UniVCD}: A New Method for Unsupervised Change Detection in the Open-Vocabulary Era}
{Zhu \MakeLowercase{\textit{et al.}}: \textit{UniVCD}: A New Method for Unsupervised Change Detection in the Open-Vocabulary Era}

\maketitle

\begin{abstract}
Change detection (CD) identifies scene changes from multi-temporal observations and is widely used in urban development and environmental monitoring. Most existing CD methods rely on supervised learning, making performance strongly dataset-dependent and incurring high annotation costs; they typically focus on a few predefined categories and generalize poorly to diverse scenes. With the rise of vision foundation models such as SAM2 and CLIP, new opportunities have emerged to relax these constraints.

We propose Unified Open-Vocabulary Change Detection (\textit{UniVCD}), an unsupervised, open-vocabulary change detection method built on frozen SAM2 and CLIP. \textit{UniVCD} detects category-agnostic changes across diverse scenes and imaging geometries without any labeled data or paired change images. A lightweight feature alignment module is introduced to bridge the spatially detailed representations from SAM2 and the semantic priors from CLIP, enabling high-resolution, semantically aware change estimation while keeping the number of trainable parameters small. On top of this, a streamlined post-processing pipeline is further introduced to suppress noise and pseudo-changes, improving the detection accuracy for objects with well-defined boundaries.
Experiments on several public BCD (Binary Change Detection) and SCD (Semantic Change Detection) benchmarks show that \textit{UniVCD} achieves consistently strong performance and matches or surpasses existing open-vocabulary CD methods in key metrics such as F1 and IoU. The results demonstrate that unsupervised change detection with frozen vision foundation models and lightweight multi-modal alignment is a practical and effective paradigm for open-vocabulary CD. Code and pretrained models will be released at \href{https://github.com/Die-Xie/UniVCD}{https://github.com/Die-Xie/UniVCD}.
\end{abstract}


\section{Introduction}

\IEEEPARstart{C}{hange} detection (CD) identifies changes by comparing multi-temporal observations of the same scene and is widely used in urban development and environmental monitoring.
Existing CD methods predominantly rely on hand-crafted deep network architectures or feature designs\cite{HANet,changeformer,changeclip}. 
Although effective on specific benchmarks, they often underperform in complex scenes and under diverse change patterns. 
Most approaches are supervised or semi-supervised and require substantial annotated data for training. 
This reliance on labels limits practical deployment, increases labeling cost and time, and makes performance strongly dependent on dataset quality. 
Moreover, these models typically exhibit poor generalization: they are often tailored to specific change types in specific scenarios and thus struggle to adapt to diverse application settings.

In recent years, vision foundation models and vision–language models (VLMs) have progressed rapidly. 
Trained on large-scale datasets, these models accumulate rich visual priors and semantic knowledge, enabling robust segmentation and recognition capabilities that can be transferred to downstream tasks. 
Representative examples include SAM2\cite{sam2} for image segmentation and CLIP\cite{clip} for open-vocabulary visual recognition. 
Motivated by these advances, it has become increasingly feasible to design more flexible and practically useful open-vocabulary change detection methods.

However, current open-vocabulary CD approaches often either invoke foundation models in a cascaded manner or directly compare high-dimensional features from these models in a coarse way. 
Such designs tend to be unstable, and detection accuracy in different scenarios is highly sensitive to the choice of backbone models and their combination strategies. 
In addition, directly applying large foundation models typically requires substantial computational resources, which further constrains their practical applicability. 
Moreover, many existing methods rely on pre-defined segmentation masks or query points for pairwise comparison and analysis\cite{segmentanychange, dynamicearth}. 
This imposes strong constraints on the shape and extent of target objects and makes it difficult to capture changes outside the given masks or query points, leading to incomplete detection. 
When imaging conditions vary significantly, these methods often struggle to remain reliable, which further restricts their applicability.

Motivated by recent progress in open-vocabulary segmentation\cite{openvocasam,ovsrs,AerOSeg}, we propose \textit{UniVCD}, an open-vocabulary unsupervised change detection method built upon CLIP and SAM2. 
\textit{UniVCD} leverages the priors learned by foundation models from large-scale data and does not require any annotated change labels. 
Through a tailored network architecture and training strategy, we fuse the semantic representations provided by CLIP with the spatially detailed features extracted by SAM2, endowing the model with high-resolution spatial–semantic representation capability and enabling open-vocabulary change detection. 
Compared with conventional CD methods, \textit{UniVCD} can effectively handle small objects and exhibits reduced sensitivity to variations in imaging viewpoint, thereby improving its suitability for real-world change detection scenarios.

The main contributions of this work are summarized as follows:
\begin{enumerate}
  \item We propose an unsupervised, open-vocabulary change detection method based on frozen SAM2 and CLIP, which performs category-agnostic change detection under diverse scenes without any labeled change data, providing a new paradigm for open-vocabulary CD.
  \item We design an efficient deep network that employs a lightweight unsupervised transfer module to align and fuse multi-scale features from the SAM2 encoder with semantic features extracted by CLIP. This multi-modal alignment preserves high-resolution spatial detail while injecting open-vocabulary semantics, substantially enhancing change detection performance with only a small number of trainable parameters.
  \item We develop a practical post-processing pipeline that combines thresholding, small-region filtering, and SAM2-based refinement to suppress noise and pseudo-changes and to sharpen object boundaries. Experiments show that this pipeline yields notable gains in precision and mean IoU for categories with clear boundaries, and we analyze its applicability across different object types.
\end{enumerate}

\begin{figure*}[!t]
	\centering
	\includegraphics[width=1.\textwidth]{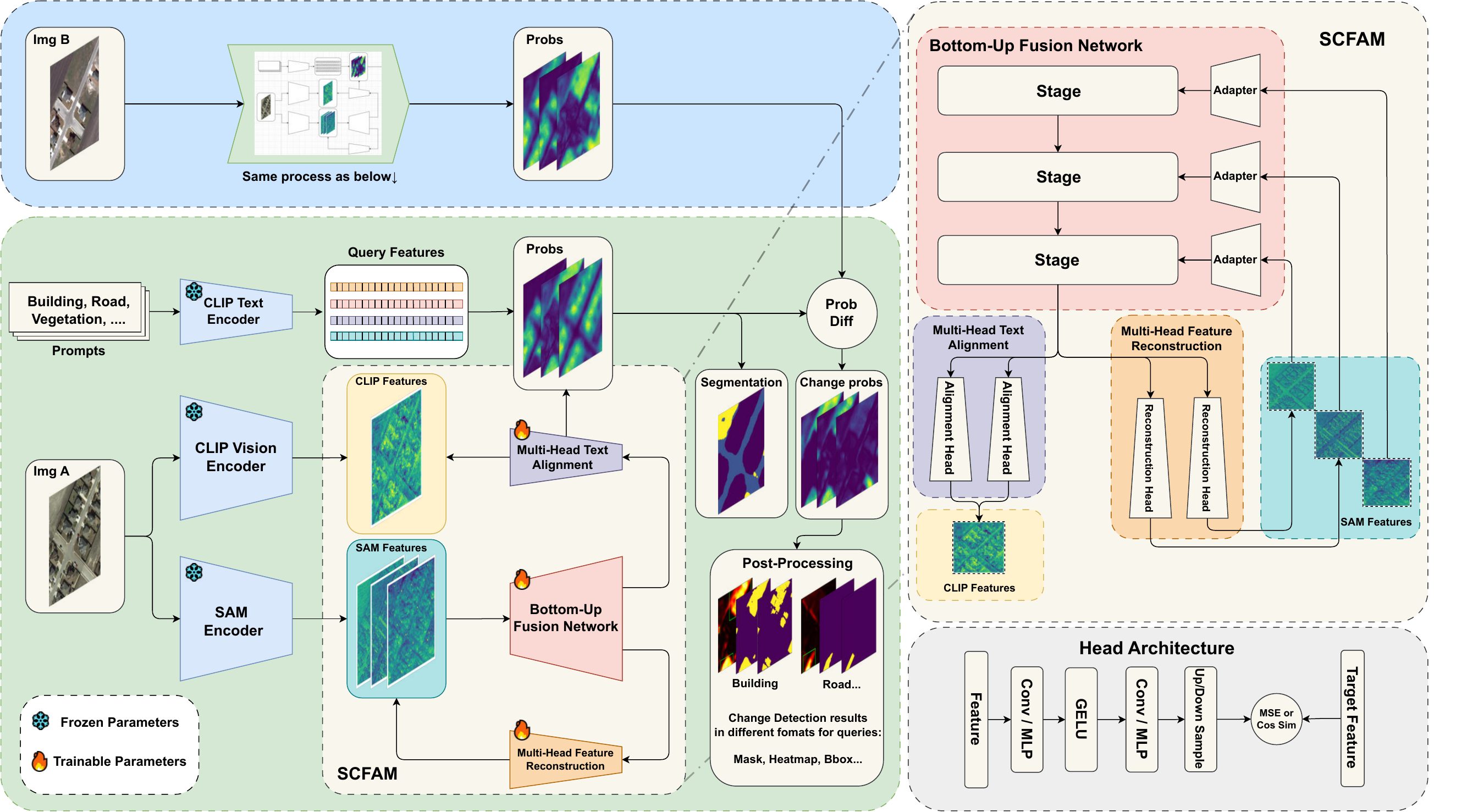}
	
	\caption{
The overall framework of the proposed \textit{UniVCD} model. The bottom-left panel illustrates the main pipeline, which consists of four components: a SAM2 encoder, a CLIP encoder, a SCFAM, and a post-processing module. The top-left panel depicts an identical workflow applied to the other image in the bi-temporal image pair. The top-right panel presents the detailed architecture of SCFAM, which mainly comprises a hierarchical feature fusion module, several adapter modules, and multiple projection heads. The bottom-right panel shows the schematic structure of a typical projection head used in our method.
		}
	\label{fig:overview}
\end{figure*}

\section{Related Work}

\subsection{Vision Foundation Models}

With the rapid advancement of deep learning, vision foundation models have achieved remarkable progress in computer vision. Trained on large-scale datasets, these models can be applied to a variety of core visual tasks, such as image classification, object detection, and semantic segmentation\cite{vit,sam2,dino,dinov2}. They accumulate rich visual priors and semantic knowledge, and thus provide strong support for a wide range of downstream applications. 

At the same time, the development of multimodal learning has endowed many vision foundation models\cite{clip,qwenvl,cogagent,florence} with cross-modal understanding capabilities. By associating visual and textual information, they can handle more complex tasks such as image captioning and visual question answering, further expanding the application scope of foundation models. 
Among purely visual foundation models, SAM2\cite{sam2} is a powerful segmentation model with strong generalization and adaptability across diverse scenes. 
Among multimodal vision–language models, CLIP\cite{clip} jointly trains image and text encoders and can effectively link visual content with semantic concepts. 
SAM3\cite{sam3} further extends these capabilities by enabling exhaustive segmentation of all instances of an open-vocabulary concept specified by a short text phrase or exemplars.

\subsection{Open-Vocabulary Change Detection Methods}

Open-vocabulary change detection aims to identify changes for arbitrary object categories without being restricted to a predefined label set. At present, research in this area is still in its early stage, with only a limited number of related studies. Existing open-vocabulary change detection methods can be broadly categorized into approaches that build directly on vision foundation models and approaches that rely on hand-crafted feature comparison.

SCM\cite{SCM} employs SAM to segment images and extract individual object regions, then applies CLIP to perform semantic classification on these regions. By comparing the similarity between features extracted by the SAM encoder and image features produced by CLIP, SCM identifies regions where changes have occurred. 
SegmentAnyChange\cite{segmentanychange} uses the SAM encoder as a feature extractor at query points and identifies changed regions by comparing the feature similarity of corresponding query areas across two images, thereby enabling point-prompted change detection.
DynamicEarth\cite{dynamicearth} investigates several change detection strategies built on different combinations of vision foundation models and vision–language models, and proposes two intuitive yet effective approaches.

\begin{figure*}[!t]
	\centering
	\includegraphics[width=\textwidth]{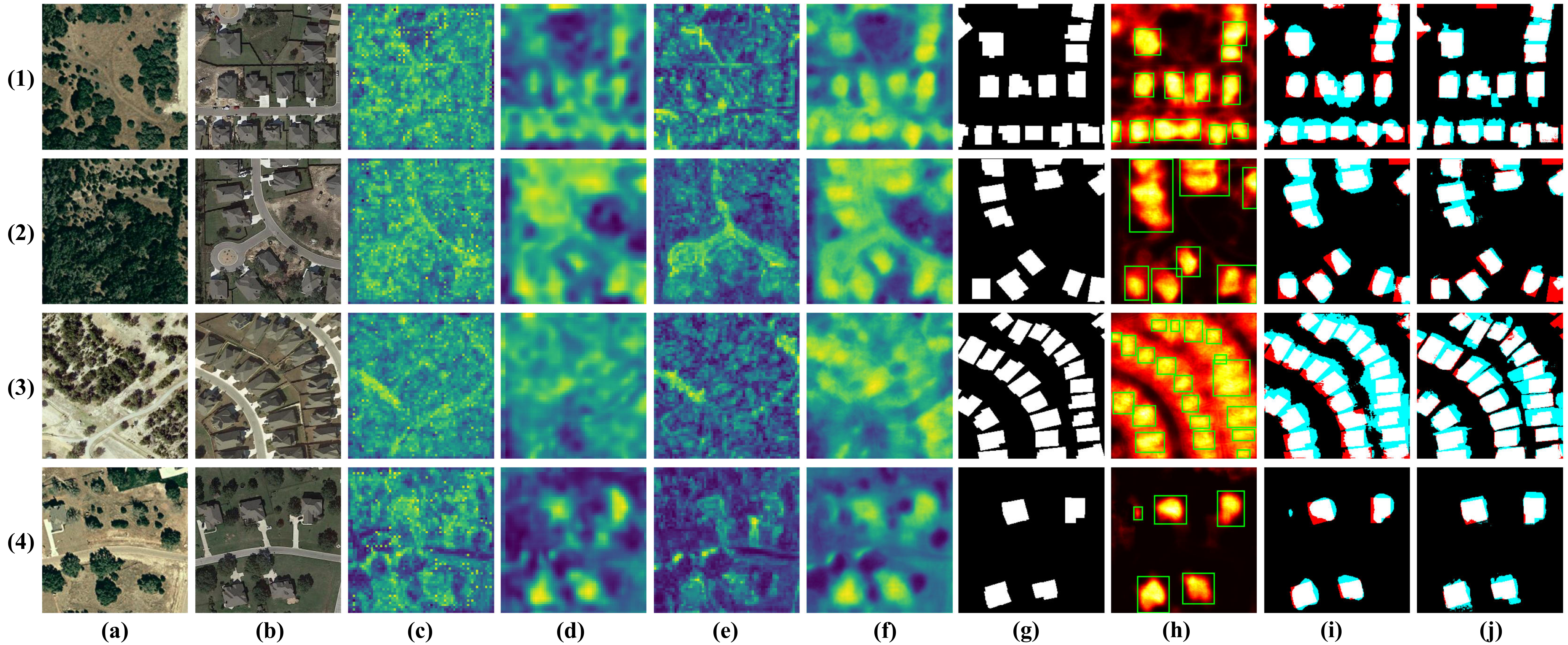}

	\caption{
Feature-map visualization and qualitative change detection results. Panels (a) and (b) show the bi-temporal remote sensing images. Panels (c) and (d) are the cosine-similarity maps of the spatial features from the SAM2 encoder and the semantic features from the CLIP encoder, respectively. Panels (e) and (f) show the fused SAM and CLIP feature maps after SCFAM. Panels (g)–(j) present the ground-truth change map, the change-probability heatmap with bounding boxes, and the binarized and post-processed change maps, respectively. In (i) and (j), white, black, red, and cyan pixels indicate true positives (TP), true negatives (TN), false positives (FP), and false negatives (FN).
		}
	\label{fig:compare_feature_maps}
\end{figure*}

\subsection{Open-Vocabulary Segmentation Methods}

Open-vocabulary segmentation aims to segment arbitrary object categories without being constrained by a predefined label set. As an important task in computer vision, it has wide practical demand, and numerous related methods have been proposed in recent years. 
Open-Vocabulary SAM\cite{openvocasam} employs transfer learning to let SAM and CLIP mutually learn each other’s feature representations, thereby enabling segmentation of arbitrary categories.
RSKT-Seg\cite{ovsrs} fuses features from DINOv2 and Remote-CLIP to effectively transfer remote sensing knowledge to open-vocabulary segmentation, improving performance on remote sensing image segmentation tasks.
SAM-CLIP\cite{samclip} jointly trains SAM and CLIP to achieve segmentation for arbitrary object categories.
OVSeg\cite{ovseg} introduces a multimodal alignment module to align visual and textual features, thus supporting open-vocabulary image segmentation.
AerOSeg\cite{AerOSeg} uses the SAM image encoder as a guidance module and trains CLIP together with a dedicated feature extraction and fusion module to perform segmentation of arbitrary categories.

\section{Methodology}
\label{sec:headings}

In this section, we focus on the concrete implementation of the proposed \textit{UniVCD}, including the overall architecture, training strategy, and post-processing pipeline, and describe the corresponding functionalities and implementation details.

\subsection{Overall Framework}
\label{sec:overall_framework}

The overall architecture of the proposed \textit{UniVCD} framework is illustrated in Fig.~\ref{fig:overview}. \textit{UniVCD} is composed of four main components: a frozen SAM2 encoder, a frozen CLIP encoder (including both image and text encoders), a SAM-CLIP Feature Alignment Module (SCFAM) that fuses multi-scale spatial and semantic features, and a post-processing module that produces the final predictions.

Both the SAM2 encoder and the CLIP encoder are kept completely frozen and are used solely for feature extraction, so as to preserve the priors and representation capacity learned from large-scale pretraining. From SAM2, we extract three spatial feature maps at different resolutions. For CLIP, we obtain image features for the bi-temporal image pair using a sliding-window strategy, and derive text features by feeding the target change categories into the text encoder with designed prompt templates.

The SCFAM module employs a lightweight unsupervised transfer mechanism to align and fuse the spatial features from SAM2 with the semantic features from CLIP. Multiple projection heads are introduced to enhance the accuracy and stability of this alignment. In this way, SCFAM preserves the high-resolution spatial representation capability of SAM2 while injecting CLIP-based semantic information, thereby leveraging the priors of both foundation models and improving the overall change detection performance.

Finally, we compute the similarity between the obtained high-resolution spatial-semantic features and the CLIP text features to derive a class-probability distribution for each pixel. These probability maps are then passed through the proposed post-processing pipeline to produce the final change detection maps and segmentation results.

\subsection{SAM2 Encoder and CLIP Encoder}

SAM2\cite{sam2} is a powerful image segmentation model that exhibits strong generality and adaptability, and can produce high-quality segmentation results across a wide range of scenes. Its encoder adopts a multi-level Transformer architecture, which can effectively capture spatial texture information and geometric boundary details in images.
To balance computational efficiency and feature representation capacity, we adopt \textit{sam2\_hiera\_large} as the backbone of the SAM2 encoder and initialize it with the official pretrained weights.
It is worth noting that the SAM2 encoder is not the only possible choice for this component in \textit{UniVCD}; other vision foundation models with strong feature extraction capability (such as SAM\cite{sam}, DINO\cite{dino}, and DINOv2\cite{dinov2}) can also serve as alternatives. However, given the requirements of change detection on small-size images, we choose SAM2 as the base encoder because of its stronger ability to capture fine-grained details.

We feed the image whose features are to be extracted (by default resized to $256 \times 256$ pixels) into the SAM2 encoder and obtain three spatial feature maps at different scales, corresponding to different levels of representation within the SAM2 encoder.

CLIP\cite{clip}, which is jointly trained on paired image and text data, has strong cross-modal understanding capability and can effectively associate visual content with semantic information. In this work, we adopt the CLIP framework provided by the ClearCLIP\cite{clearclip} project. For image encoding, we extract features in a block-wise manner using a sliding-window strategy, together with an appropriate overlap ratio and smoothing scheme to maintain the continuity and completeness of the extracted features and to reduce discontinuities or information loss at window boundaries. 
For text encoding, we use the CLIP text encoder to encode the target change categories and construct multiple prompt templates to generate the corresponding textual feature representations, thereby improving the robustness, and stability. 
Similar to the SAM2 encoder, the CLIP encoder can also be replaced by other CLIP variants or pretrained models (such as RemoteCLIP\cite{remoteclip} and BLIP\cite{blip}), thereby further improving the adaptability of the proposed framework to specific application scenarios.

\subsection{SAM-CLIP Feature Alignment Module (SCFAM)}
\label{sec:scfam}

Figure~\ref{fig:overview}(top-right) illustrates the detailed architecture of the SCFAM module. It mainly consists of a layer-wise feature fusion block, several adapter modules, and multiple projection heads. The primary goal of SCFAM is to align and fuse the multi-scale spatial features extracted by the SAM2 encoder with the semantic features produced by the CLIP encoder.

More specifically, SCFAM first feeds the three spatial feature maps at different scales extracted by the SAM2 encoder into a layer-wise feature fusion module. Each feature map is initially passed through a residual adapter module to align the channel dimensions. The fusion module then progressively aggregates information from bottom to top. For the fusion backbone, we adopt a modified ConvNeXt\cite{convnext} architecture to exploit its strengths in visual feature extraction, and employ upsampling and residual convolutions to ensure effective information propagation across different resolutions during the fusion process.

For the adapter module, we adopt a lightweight design to reduce computational overhead and improve training efficiency. Each adapter mainly consists of several convolution layers with small kernels, in order to meet the requirements of feature alignment across different scales. Concretely, we use a lightweight convolutional alignment block $\mathcal{F}_i(\cdot)$ for channel mapping and local context modeling. Given the input feature at the $i$-th scale $\mathbf{X}_i \in \mathbb{R}^{C_i^{\text{sam}}\times H\times W}$, we employ a two-layer convolutional alignment module $\mathcal{F}_i$ to project it into a unified semantic space, yielding
\begin{align}
\mathbf{Y}_i = \mathcal{F}_i(\mathbf{X}_i),
\end{align}
where $\mathcal{F}_i$ consists of a $1\times 1$ convolution for channel alignment followed by a $3\times 3$ convolution with normalization and nonlinear activation to model local contextual information, which can be abstractly written as
\begin{align}
\mathcal{F}_i(\mathbf{X}_i)
= \mathrm{Conv}_{3\times 3}^{(i)}\!\bigl(
    \phi\bigl(
        \mathrm{Conv}_{1\times 1}^{(i)}(\mathbf{X}_i)
    \bigr)
\bigr),
\end{align}
where $\phi(\cdot)$ denotes the standard normalization and activation operations, including LayerNorm and GELU.

Finally, for the feature maps after processing, we apply the corresponding projection heads to project and map the fused features. We design two groups of projection heads: one group is used to reconstruct the SAM2 feature maps at each scale, and the other group is used to align the fused features with the CLIP image features.
The detailed structure of the projection heads is illustrated in the bottom-right panel of Fig.~\ref{fig:overview}. Inspired by\cite{AerOSeg,clipzeroseg}, each projection head mainly consists of several MLPs or convolution layers with small kernels, together with nonlinear activation functions. This design allows the projection heads to effectively project and map the input features into the desired target feature spaces.

\begin{figure*}[!t]
	\centering
	\includegraphics[width=\textwidth]{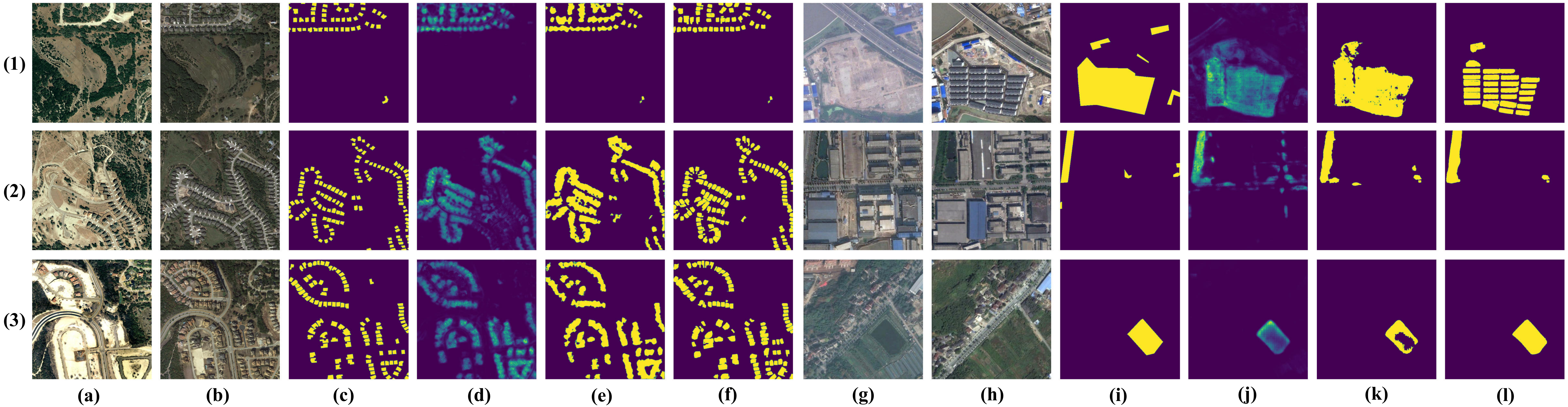}

	\caption{
High-resolution change detection results of \textit{UniVCD}. Panels (a)–(f) have a resolution of $1024 \times 1024$, and panels (g)–(l) have a resolution of $512 \times 512$. For each group, (a)/(g) and (b)/(h) are the bi-temporal images, (c)/(i) are the ground-truth change maps, (d)/(j) are the change-likelihood heatmaps, and (e)/(k) and (f)/(l) show the binarized and post-processed change maps.
		}
	\label{fig:large_img}
\end{figure*}

\subsection{Loss Design}
\label{sec:loss_design}

To ensure that \textit{UniVCD} can effectively align and fuse features at different scales, we design a multi-scale alignment loss, which consists of the following components:

\begin{enumerate}
  \item \textbf{Reconstruction loss:} For each scale of the SAM2 feature maps, we use the mean squared error (MSE) loss to measure the discrepancy between the projected features and the original SAM2 features. This loss is applied to the three SAM projection heads corresponding to the two spatial scales.
  \item \textbf{Alignment loss:} For the CLIP image features, we use a Mean Cosine Similarity (MCS) loss to enforce global directional alignment in feature space and an MSE loss to reduce local discrepancies in feature values. These two losses jointly encourage consistent alignment at both global and local levels and are applied to the two CLIP projection heads.
\end{enumerate}

The specific formulations of the MSE and Mean Cosine Similarity losses are given by
\begin{align}
\mathcal{L}_{\text{MSE}}(\mathbf{A}, \mathbf{B}) &= \frac{1}{N} \sum_{i=1}^{N} \bigl(\mathbf{A}_i - \mathbf{B}_i\bigr)^2, \\
\mathcal{L}_{\text{MCS}}(\mathbf{A}, \mathbf{B}) &= 1 - \frac{1}{N} \sum_{i=1}^{N} \frac{\mathbf{A}_i \cdot \mathbf{B}_i}{\|\mathbf{A}_i\| \,\|\mathbf{B}_i\|},
\end{align}
where $\mathbf{A}_i$ and $\mathbf{B}_i$ denote the feature vectors at position $i$ in the two feature sets, and $N$ is the total number of positions.

Due to the potentially large differences in magnitude among these loss terms during training, we introduce weighting coefficients $\lambda_i$ to balance their contributions. The overall loss function is defined as
\begin{align}
\mathcal{L}_{\text{total}} = \sum_{i=1}^{5} \lambda_i \mathcal{L}_i,
\end{align}
where $\mathcal{L}_i$ denotes the individual loss components described above and $\lambda_i$ is the corresponding weight. In practice, we set the reconstruction loss to be approximately two orders of magnitude smaller than the alignment losses.

As illustrated in Figs.~\ref{fig:compare_feature_maps}~(c)-(f), we further compare the cosine-similarity maps of features extracted from a representative bi-temporal remote sensing image pair. The spatial feature map from the SAM2 encoder mainly emphasizes texture and boundary information, whereas the semantic feature map from the CLIP encoder focuses on global semantics but exhibits low spatial resolution and block-like artifacts introduced by the sliding-window extraction, as well as noise arising from imperfect semantic understanding. In contrast, the fused feature maps produced by SCFAM successfully combine the strengths of both: they preserve high-resolution spatial details while incorporating rich semantic cues, thus providing a more informative representation for subsequent change detection.

\subsection{Post-Processing for Change Detection and Segmentation}
\label{sec:cd_seg_methods}

For open-vocabulary segmentation, we follow the CLIP\cite{clip} paradigm and compute the cosine similarity between the fused feature map and the CLIP text features to obtain a per-pixel class probability map. The process can be formulated as
\begin{align}
S_{i,j} = \frac{\mathbf{F}_{i,j} \cdot \mathbf{T}}{\|\mathbf{F}_{i,j}\| \,\|\mathbf{T}\|},
\end{align}
where $S_{i,j}$ denotes the similarity score (logit) at pixel $(i,j)$, $\mathbf{F}_{i,j}$ is the feature vector at pixel $(i,j)$ in the fused feature map, and $\mathbf{T}$ denotes the text feature vector.

For change detection, in order to support open-vocabulary settings, we first obtain the class-probability maps $S$ for each image as described above, yielding $S_1$ and $S_2$ for the bi-temporal pair. We then compute, for each category, the difference between the corresponding probability maps to obtain an initial estimate of the change likelihood. Specifically, we compute
\begin{align}
D_c(i,j) = \bigl(S_{1,c}(i,j) - S_{2,c}(i,j)\bigr)^2,
\end{align}
where $D_c(i,j)$ denotes the change likelihood of category $c$ at pixel location $(i,j)$, and $S_{1,c}(i,j)$ and $S_{2,c}(i,j)$ are the corresponding class-probability values for category $c$ at pixel $(i,j)$ in the first and second image, respectively.

Compared with directly computing cosine similarity between the feature maps of the two images, this formulation makes more effective use of CLIP's semantic representations and can improve the accuracy of change detection to some extent. More importantly, it yields an independent change-likelihood map for each category, enabling open-vocabulary change detection as well as joint multi-class change detection within a unified framework, while requiring only a single training procedure to support arbitrary target categories.

To obtain a binarized change map from the change-likelihood map while suppressing noise and pseudo-changes, we design the following post-processing procedure:

\begin{enumerate}
  \item We first apply the Otsu\cite{otsu} algorithm to binarize the change-likelihood map and obtain an initial change mask. Morphological opening and small-region filtering are then used to suppress noise and spurious isolated responses, from which we derive a set of preliminary candidate change regions.
  \item \label{sam2 post} Next, we refine these candidate regions using the SAM2 model. For high-confidence change areas, we generate bounding boxes and point prompts to guide SAM2-based segmentation, and use the initial candidate mask to discard low-overlap false positives. This yields more accurate and spatially precise change boundaries. 
  \item Optionally, we also experiment with directly using the prompt text as input to SAM3 to achieve even finer segmentation results.
\end{enumerate}

Through the above design, \textit{UniVCD} can perform open-vocabulary change detection in an efficient and accurate manner. Unlike conventional change detection models that directly output binary masks in an end-to-end fashion, \textit{UniVCD} produces probabilistic change maps, which offer higher flexibility and interpretability. Users can adjust thresholds or modify the post-processing steps according to specific requirements, and can also export the results in different formats (e.g., heatmaps or bounding boxes) to accommodate diverse application scenarios and change detection needs.

However, it should be noted that, due to substantial variations across images, the post-processing step may increase detection precision at the expense of a certain decrease in recall. Therefore, we recommend applying the SAM2-based refinement in Step~\ref{sam2 post} to scenarios with clear object boundaries and pronounced changes (e.g., buildings and roads) to further improve precision, while for scenarios with ambiguous boundaries or subtle changes (e.g., vegetation and soil), this post-processing may not yield significant improvements.

\label{sec:post_limitations}
In addition, since \textit{UniVCD} performs category-wise detection, when a region changes but its semantic category remains unchanged, \textit{UniVCD} may not capture such changes as expected.

\begin{figure}[t]
	\centering
	\includegraphics[width=\columnwidth]{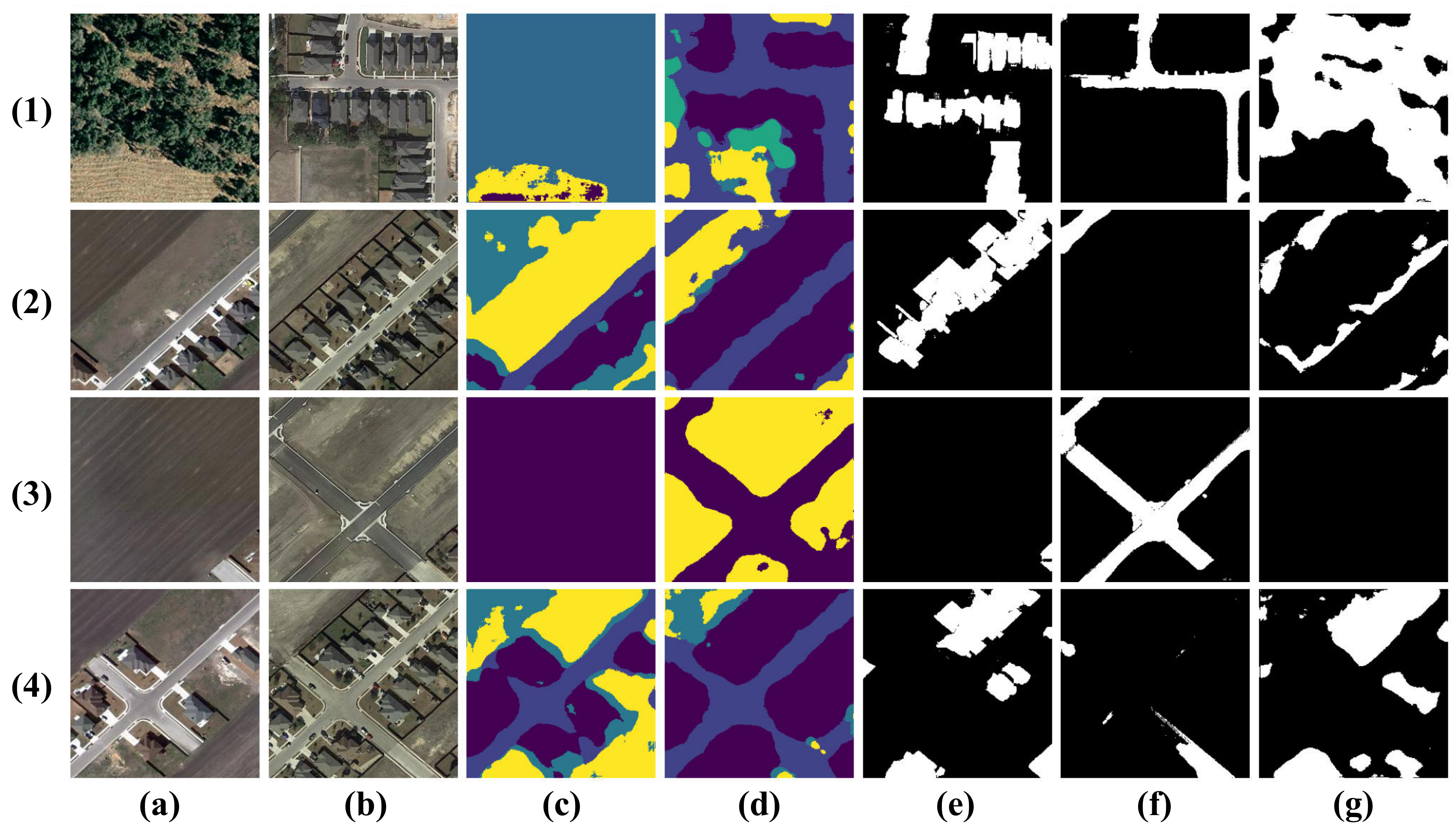}
	
	\caption{
Open-vocabulary capabilities of \textit{UniVCD} in segmentation and change detection. Panels (a) and (b) show the bi-temporal remote sensing images; (c) and (d) give the corresponding open-vocabulary segmentation maps, where colors only distinguish categories within each image. Panels (e)–(g) show the change detection results for \textit{Building}, \textit{Road}, and \textit{Vegetation}, respectively.
		}
	\label{fig:multi_cd_seg}
\end{figure}

Figures~\ref{fig:compare_feature_maps}(g)-(j) show the post-processing results for the \textit{Building} category on images with a resolution of $256 \times 256$, while Fig.~\ref{fig:multi_cd_seg} presents representative results on open-vocabulary change detection and segmentation. For objects with well-defined boundaries, the proposed post-processing pipeline can improve detection accuracy and yield clearer, more precise change boundaries.

To handle inputs at different spatial resolutions, we adopt an overlapped sliding-window strategy for high-resolution images: the image is divided into overlapping patches, features and change maps are computed patch by patch, and the patch-wise results are then stitched and fused to obtain the final full-image change map. Figure~\ref{fig:large_img} illustrates multi-class change detection results on high-resolution remote sensing images. Panels (a)-(f) are from the LEVIR-CD\cite{LEVIR} dataset with a resolution of $1024 \times 1024$ and focus on building changes, whereas panels (g)-(l) are from the SECOND\cite{SECOND} dataset with a resolution of $512 \times 512$, where (1)-(3) correspond to changes in \textit{building}, \textit{tree}, and \textit{water}, respectively. The results indicate that \textit{UniVCD} can effectively process high-resolution images and detect changes for multiple object categories.

\section{Experiment}

\begin{table*} 
	
	\caption{Comparison of change detection results on the BCD datasets. The top three scores are highlighted in \textcolor{red}{RED}, \textcolor{purple}{PURPLE}, and \textcolor{brown}{BROWN}, respectively. A dash “-” indicates missing results. $P^c$, $R^c$, $F1^c$, and $IOU^c$ denote precision, recall, F1 score, and IOU for the changed class, respectively, and $mIOU$ denotes the mean IOU over all classes.}

	\centering

  	\renewcommand{\arraystretch}{1.15}
	\begin{tabular}{l|ccccc|ccccc}
		
    	\noalign{\vskip 6pt}
		\hline
		\multirow{2}{*}{Method} & \multicolumn{5}{c|}{LEVIR-CD} 
								& \multicolumn{5}{c}{WHU-CD}\\

     	& $P^c$ & $R^c$ & $F1^c$ & $IOU^c$ & $mIOU$ 
		& $P^c$ & $R^c$ & $F1^c$ & $IOU^c$ & $mIOU$
		\\
		\hline
		PCA-KM\cite{PCA-KM} & - & - & 9.1 & 4.8 & 28.9 
							& - & - & 10.2 & 5.4 & 31.7
							\\
		CNN-CD\cite{CNN-CD} & - & - & 13.1 & 7.0 & 35.4 
							& - & - & 9.4 & 4.9 & 33.7
							\\
		DSFA\cite{DSFA}     & - & - & 8.2 & 4.3 & 40.7 
							& - & - & 7.8 & 4.1 & 42.5
							\\
		DCVA\cite{DCVA} 	& - & - & 14.1 & 7.6 & 46.1 
							& - & - & 19.6 & 10.9 & 44.9
							\\
		GMCD\cite{GMCD} 	& - & - & 11.6 & 6.1 & 42.8 
							& - & - & 19.7 & 10.9 & 47.6
							\\
		CVA\cite{CVA} 		& 7.5 & 32.6 & 12.2 & - 
							& - & - & - & - & - & -
							\\
		DINOv2+CVA(ViT-G)\cite{segmentanychange} 	& 9.5 & \textcolor{red}{96.6} & 17.3 & - & - 
													& - & - & - & - & -
													\\
		SAM+Mask Match(ViT-H)\cite{segmentanychange}& 12.6 & 30.2 & 17.8 & - & - 
													& - & - & - & - & -
													\\
		AnyChange\cite{segmentanychange} 			& 13.3 & \textcolor{brown}{85.0} & 23.0 & - & - 
													& - & - & - & - & -
													\\
		SCM\cite{SCM} 								& - & - & 31.7 & 18.8 & 53.7 
													& - & - & 31.3 & 18.6 & 52.1
													\\
		SAM-DINO-SegEarth-OV\cite{dynamicearth} 	& - & - & 49.7 & 33.0 & - 
													& - & - & 53.7 & 36.7 & -
													\\
		SAM-DINOv2-SegEarth-OV\cite{dynamicearth} 	& - & - & 53.6 & 36.6 & - 
													& - & - & 57.7 & 40.6 & -
													\\
		SAM2-DINOv2-SegEarth-OV\cite{dynamicearth} 	& - & - & 50.5 & 33.8 & - 
													& - & - & 58.1 & 40.9 & -
		\\
													APE-/-DINO\cite{dynamicearth} 				& - & - & \textcolor{brown}{69.7} & \textcolor{brown}{53.5} & - 
													& - & - & 72.5 & 56.8 & -
													\\
		APE-/-DINOv2\cite{dynamicearth} 			& - & - & 66.7 & 50.0 & - 
													& - & - & \textcolor{brown}{75.8} & \textcolor{brown}{61.1} & -
													\\

		SAM3-CD 			& 42.6 & \textcolor{purple}{86.9} & 57.1 & 40.0 & 66.5 
							& 36.9 & 95.3 & 53.2 & 36.3 & 63.6
							\\
		\hline
		\textit{UniVCD}(crop.) 			& 47.6 & 83.0 & 60.5 & 43.4 & 68.8 
									& 37.1 & 72.9 & 49.2 & 32.6 & 63.2
									\\
		\textit{UniVCD}(crop., postproc) 	& \textcolor{brown}{58.1} & 77.4 & 66.4 & 49.7 &\textcolor{brown}{72.8}
									& 57.0 & 71.4 & 63.4 & 46.4 & 71.5
									\\
		\textit{UniVCD}(orig.) 			& 49.7 & 81.3 & 61.7 & 44.6 & 69.6
									& 62.0 & 77.7 & 69.0 & 52.6 & \textcolor{purple}{74.5}
									\\
		\textit{UniVCD}(orig., postproc) 	& \textcolor{purple}{64.7} & 77.9 & \textcolor{purple}{70.7} & \textcolor{purple}{54.7} & \textcolor{purple}{75.6}
									& 70.2 & 84.1 & \textcolor{purple}{76.5} & \textcolor{purple}{61.9} & \textcolor{purple}{79.6}
									\\

		\textit{UniVCD}(orig., SAM3 postproc) 	& \textcolor{red}{65.1} & 81.4 & \textcolor{red}{72.3} & \textcolor{red}{56.7} & \textcolor{red}{76.7}
									& 73.3 & 86.8 & \textcolor{red}{79.5} & \textcolor{red}{66.0} & \textcolor{red}{81.2}
									\\
		\hline
	\end{tabular}
	\label{tab:bcd_results}
\end{table*}

\begin{table*}

	\caption{Comparison of change detection results on the SECOND dataset. The top three scores are highlighted in \textcolor{red}{RED}, \textcolor{purple}{PURPLE}, and \textcolor{brown}{BROWN}, respectively. $F1^c$ and $IOU^c$ denote the F1 score and IOU of the changed class.}
	\centering

  	\renewcommand{\arraystretch}{1.15}
	\begin{tabular}{l|cc|cc|cc|cc|cc|cc}
		
    	\noalign{\vskip 6pt}
		\hline
    	Method & \multicolumn{2}{c|}{Building} & \multicolumn{2}{c|}{Tree}
           	& \multicolumn{2}{c|}{Water} & \multicolumn{2}{c|}{Low Veg.}
           	& \multicolumn{2}{c|}{N.V.g surface} & \multicolumn{2}{c}{Playground} \\

           	& $F1^c$ & $IOU^c$ & $F1^c$ & $IOU^c$ & $F1^c$ & $IOU^c$
           	& $F1^c$ & $IOU^c$ & $F1^c$ & $IOU^c$ & $F1^c$ & $IOU^c$ \\
    	\hline
		SAM-DINO-SegEarth-OV\cite{dynamicearth}  & 50.8 & 34.1  
													& 28.3 & 16.5  
													& 23.6 & 13.4  
													& \textcolor{brown}{38.7} & \textcolor{brown}{24.0}  
													& 36.7 & 22.5  
													& 27.6 & 16.0  
		\\
		SAM-DINOv2-SegEarth-OV\cite{dynamicearth} & 55.2 & 38.1  
													& \textcolor{red}{33.8} & \textcolor{red}{20.3}  
													& \textcolor{purple}{25.1} & \textcolor{purple}{14.3}  
													& \textcolor{purple}{38.9} & \textcolor{purple}{24.1}  
													& 41.6 & 26.2  
													& \textcolor{red}{33.3} & \textcolor{red}{20.0}  
		\\
		SAM2-DINOv2-SegEarth-OV\cite{dynamicearth} & 53.5 & 36.6  
													& 30.8 & 18.2  
													& \textcolor{brown}{24.3} & \textcolor{brown}{13.8}  
													& 36.2 & 22.1  
													& 32.3 & 19.2  
													& \textcolor{purple}{29.2} & \textcolor{purple}{17.1}  
		\\
		APE-/-DINO\cite{dynamicearth}             & 42.0 & 26.5  
													& 23.8 & 13.5  
													& 17.9 & 9.8   
													& $\approx 0$ & $\approx 0$      
													& $\approx 0$ & $\approx 0$     
													& 28.3 & 16.5  
		\\
		APE-/-DINOv2\cite{dynamicearth}           & 43.9 & 28.1  
													& 24.8 & 14.1  
													& 21.7 & 12.2  
													& 2.7  & 1.4   
													& $\approx 0$ & $\approx 0$     
													& 27.6 & 16.0  
		\\
		SAM3-CD           							& 36.9 & 22.6  
													& 22.1 & 12.4  
													& 9.6 & 5.0  
													& 30.9 & 18.2  
													& 42.1 & 26.7  
													& 21.1 & 11.8  
		\\
		\hline

		\textit{UniVCD}(orig.) 			& 55.2 & 38.1
											& \textcolor{brown}{32.4} & \textcolor{brown}{19.3}
											& 6.2 & 3.2
											& 35.0 & 21.2
											& \textcolor{purple}{43.3} & \textcolor{purple}{27.7}
											& $\approx 0$ & $\approx 0$
		\\
		\textit{UniVCD}(orig., postproc) 	& \textcolor{purple}{58.4} & \textcolor{purple}{41.2}
												& \textcolor{purple}{32.5} & \textcolor{purple}{19.4}
												& 5.0 & 2.6
												& 37.6 & 23.2
												& \textcolor{brown}{42.6} & \textcolor{brown}{27.1}
												& $\approx 0$ & $\approx 0$
		\\
		\textit{UniVCD}(orig., SAM3 postproc) 	& \textcolor{red}{60.4} & \textcolor{red}{43.2}
												& 31.9 & 18.9
												& 15.2 & 8.2
												& \textcolor{red}{39.8} & \textcolor{red}{24.9}
												& \textcolor{red}{43.7} & \textcolor{red}{28.0}
												& $\approx 0$ & $\approx 0$
		\\

		\textit{UniVCD}-RemoteClip(orig., postproc) 	& \textcolor{brown}{56.6} & \textcolor{brown}{39.5}
														& 26.3 & 15.1
														& \textcolor{red}{27.4} & \textcolor{red}{15.8}
														& 33.2 & 19.9
														& 26.9 & 15.5
														& $\approx 0$ & $\approx 0$
		\\
		\hline
	\end{tabular}
	\label{tab:scd_results}
\end{table*}

\subsection{Implementation Details}

The proposed \textit{UniVCD} model is implemented in the PyTorch framework, and all training and testing can be carried out on a single NVIDIA RTX 4090 GPU. We use the AdamW optimizer with a learning rate of $1\times10^{-4}$ and a weight decay of $1\times10^{-5}$. No data augmentation is applied to the input images. During training, only unpaired images are used as inputs for unsupervised learning, with a default batch size of 3. The total number of training epochs is set to 10-60 to avoid the model overfitting to either semantic or spatial features.

For the BCD datasets, we use the following prompt words for change detection:
\textit{"architecture"},
\textit{"road"},
\textit{"vegetation"},
\textit{"water"},
\textit{"bare ground"}.

For the SCD dataset, we further extend the original label set by adding the \textit{"road"} category, so that roads can be distinguished from buildings.

\subsection{Datasets}

To comprehensively evaluate the performance of \textit{UniVCD} on open-vocabulary change detection, we conduct experiments on several public benchmark datasets. Since \textit{UniVCD} is trained in an unsupervised manner, we only use the test sets for evaluation.

Considering computational resources and efficiency, training is performed on non-overlapping image patches of size $256 \times 256$ pixels. During testing, we evaluate the model both on the original high-resolution images (cropped to $1024 \times 1024$ pixels when necessary) and on non-overlapping $256 \times 256$ patches, in order to assess its performance under different input resolutions. The datasets are described as follows:

\begin{enumerate}
  \item \textbf{LEVIR-CD}\cite{LEVIR}: This dataset is collected from Google Earth and contains 637 pairs of high-resolution remote sensing images, of which 128 pairs are used for testing. It is primarily designed for building change detection. Each image pair has a resolution of $1024 \times 1024$ pixels with a spatial resolution of 0.5 m, and includes various types of building changes such as new construction, demolition, and expansion.
  \item \textbf{WHU-CD}\cite{WHUCD}: This dataset is derived from satellite and aerial imagery. After cropping to $256 \times 256$ pixels, it contains 7{,}434 image pairs, including 744 test pairs. We also evaluate on cropped $1024 \times 1024$ images, where the test set consists of 48 pairs. The spatial resolution is 0.075 m for aerial images and 2.7 m for satellite images. WHU-CD mainly targets building change detection in diverse and complex urban environments.
  \item \textbf{SECOND}\cite{SECOND}: This dataset comprises 4{,}662 pairs of $512 \times 512$ multi-platform remote sensing images. Since no official split is provided, we adopt the partitioning protocol of SCanNet\cite{SCanNet}, using 593 pairs for testing. SECOND covers six land-cover change types: non-vegetated ground surface (N.V.g surface), tree, low vegetation, water, building, and playground.
\end{enumerate}

\subsection{Evaluation Metrics}

We adopt commonly used evaluation metrics for change detection, including Precision, Recall, F1 Score, Intersection over Union (IOU), and mean IOU (mIOU), to comprehensively assess the performance of the proposed method. The metrics are defined as follows:

\begin{align}
\mathrm{Precision} &= \frac{TP}{TP + FP},\\
\mathrm{Recall} &= \frac{TP}{TP + FN},\\
\mathrm{F1} &= 2 \cdot \frac{\mathrm{Precision} \cdot \mathrm{Recall}}{\mathrm{Precision} + \mathrm{Recall}},\\
\mathrm{IoU} &= \frac{TP}{TP + FP + FN},\\
\mathrm{mIoU} &= \frac{1}{C} \sum_{c=1}^{C} \frac{TP_c}{TP_c + FP_c + FN_c}.
\end{align}

where $TP$, $FP$, and $FN$ denote the numbers of true positives, false positives, and false negatives, respectively; $C$ is the total number of classes; and $TP_c$, $FP_c$, and $FN_c$ denote the numbers of true positives, false positives, and false negatives for class $c$, respectively.

\begin{table*} 
	\caption{Ablation study of the proposed SCFAM module on LEVIR-CD and WHU-CD datasets using original-resolution images for testing.}
    \centering

      \renewcommand{\arraystretch}{1.15}
    \begin{tabular}{l|c|ccccc|ccccc}
        \noalign{\vskip 6pt}
        \hline
        \multirow{2}{*}{Method} &
        \multirow{2}{*}{Postproc} &
        \multicolumn{5}{c|}{LEVIR-CD} &
        \multicolumn{5}{c}{WHU-CD} \\
        & &
        $P^c$ & $R^c$ & $F1^c$ & $IOU^c$ & $mIOU$ &
        $P^c$ & $R^c$ & $F1^c$ & $IOU^c$ & $mIOU$ \\
        \hline

        \textit{UniVCD} w/o SCFAM	  &   & 43.6 & 70.5 & 53.9 & 36.9 & 65.2 
                                              & 44.1 & 61.9 & 51.5 & 34.7 & 64.3 \\
        \textit{UniVCD} w/o Recon. & No & 46.3 & 76.2 & 57.6 & 40.4 & 67.2
                                              & 55.0 & 67.3 & 60.5 & 43.4 & 69.4 \\
        \textit{UniVCD}            &  & 49.7 & 81.3 & 61.7 & 44.6 & 69.6
                                              & 62.0 & 77.7 & 69.0 & 52.6 & 74.5 \\
        \hline
        
        \textit{UniVCD} w/o SCFAM            &  & 49.2 & 74.5 & 59.3 & 42.1 & 68.4 
                                              & 50.2 & 67.9 & 57.7 & 40.5 & 67.6 \\
        \textit{UniVCD} w/o Recon. & Yes & 51.2 & 78.3 & 62.3 & 45.2 & 70.0
                                              & 62.8 & 69.9 & 66.1 & 49.4 & 72.8 \\
        \textit{UniVCD}            &  & 64.7 & 77.9 & 70.7 & 54.7 & 75.6
                                              & 70.2 & 84.1 & 76.5 & 61.9 & 79.6 \\
        \hline
    \end{tabular}
    \label{tab:ablation_study}
\end{table*}

\subsection{Experimental Results}

As shown in Table~\ref{tab:bcd_results}, we compare \textit{UniVCD} with a range of classical methods and recent unsupervised approaches on the BCD datasets. To assess performance under different input resolutions, we evaluate the model both on cropped $256 \times 256$ patches (crop.) and on the original high-resolution images (orig.). 
Moreover, to evaluate the impact of the post-processing step described in Section~\ref{sam2 post}, we report results with and without post-processing (postproc), as well as results using the SAM3-based post-processing (sam3 postproc).

Meanwhile, we implement SAM3-CD, a baseline open-vocabulary change detection method directly based on SAM3. The specific implementation details can be found in Appendix~\ref{appendix:sam3cd}.

As can be seen, the proposed \textit{UniVCD} model performs well under both resolutions. In particular, applying the post-processing step substantially improves F1 and IOU scores, achieving state-of-the-art performance.

Moreover, \textit{UniVCD} attains better results on the original high-resolution images than on the cropped patches, indicating that the model can effectively exploit high-resolution information for change detection. We attribute this gain to two factors: on the one hand, high-resolution inputs avoid the loss of boundary details and provide more complete contextual information compared with patch-wise cropping; on the other hand, they reduce the risk that SAM2 over-segments image textures due to overly small input sizes in the post-processing stage, thereby leading to more accurate change detection.

As shown in Table~\ref{tab:scd_results}, we report the change detection performance of \textit{UniVCD} on the SECOND dataset. \textit{UniVCD} achieves strong results across multiple categories and performs favorably on various change types. In particular, it clearly outperforms competing methods on categories such as \textit{building} and \textit{tree}, highlighting its effectiveness on complex change detection tasks. Consistent with the discussion in Section~\ref{sec:cd_seg_methods}, the post-processing step brings notable gains for classes with clear object boundaries (e.g., buildings), whereas its benefit is more limited for classes with fuzzy boundaries (e.g., low vegetation).

At the same time, we observe that \textit{UniVCD} performs relatively weakly on certain categories (e.g., water). We attribute this mainly to the fact that the adopted CLIP\cite{clearclip} model is not specifically pretrained on remote sensing imagery, which limits its semantic discrimination ability for some domain-specific categories and thus degrades change detection performance. To verify this hypothesis, we replace the original CLIP encoder with RemoteCLIP\cite{remoteclip} (denoted as \textit{UniVCD}-RemoteCLIP). Preliminary results show that, for the water category, \textit{UniVCD}-RemoteCLIP achieves change detection performance exceeding existing state-of-the-art methods, supporting our conjecture. 

However, for the playground category, we believe that the limitation mainly arises from the SAM2 encoder's insufficient spatial feature representation for certain classes. In future work, we plan to further investigate and improve this component, and to refine remote-sensing–oriented multimodal pretrained models so as to enhance the performance of \textit{UniVCD} across different categories in change detection tasks.

Across multiple datasets, \textit{UniVCD} exhibits consistently strong and stable performance, demonstrating its effectiveness and broad applicability for open-vocabulary change detection. The model can handle images with different spatial resolutions and further benefits from the proposed post-processing step, which enhances detection performance and highlights its potential for practical deployment.

Meanwhile, \textit{UniVCD} is also applicable to change detection for small objects and for oblique photography, which are capabilities that many existing methods lack. Since current datasets rarely focus on these scenarios, we provide qualitative results in Appendix~\ref{sec:other_features}.

\subsection{Ablation Study}

To validate the effectiveness of the proposed SCFAM module for cross-modal feature alignment and fusion, we conduct ablation studies. Specifically, we design the following two experimental settings:

\begin{enumerate}
  \item \textbf{w/o SCFAM}: We remove the SCFAM module and perform change detection using only the features from the CLIP encoder, in order to evaluate the effectiveness of single-modal features for the change detection task.
  \item \textbf{w/o Reconstruction Loss}: We remove the reconstruction loss term in SCFAM and retain only the alignment loss, so as to assess the impact of the reconstruction loss on the overall performance.
\end{enumerate}

We conduct the experiments on the LEVIR-CD and WHU-CD test sets using $1024 \times 1024$ resolution images, with all other settings kept consistent with the previous experiments. The change detection results under different experimental settings are presented in Table~\ref{tab:ablation_study}.

As summarized in Table~\ref{tab:ablation_study}, removing the SCFAM module leads to a pronounced degradation in change detection performance, indicating that cross-modal feature alignment and fusion play a crucial role in improving the results. Likewise, discarding the reconstruction loss also causes a performance drop, which suggests that this loss helps preserve the completeness and expressiveness of the features and thus enhances detection accuracy. Overall, the SCFAM module together with its loss design is essential to the success of \textit{UniVCD} on open-vocabulary change detection tasks.

\section{Conclusion}

This paper presents \textit{UniVCD}, a large-model–based method for open-vocabulary change detection. By combining the strong feature extraction capabilities of SAM2 and CLIP and introducing an efficient cross-modal feature alignment and fusion module (SCFAM), \textit{UniVCD} enables unsupervised change detection for arbitrary object categories in remote sensing images. Experimental results on multiple public datasets show that \textit{UniVCD} achieves superior performance compared with existing methods, demonstrating its effectiveness and broad applicability in open-vocabulary change detection. In future work, we plan to further refine the model architecture and explore more efficient feature alignment representations to enhance the performance of \textit{UniVCD} across different change detection categories.

\vspace{11pt}

\bibliographystyle{IEEEtran}
\bibliography{references}{}

\appendices

\section{Implementation Details of SAM3-CD}
\label{appendix:sam3cd}

The overall implementation of SAM3-CD is relatively concise and straightforward. 
Given a pair of temporal images $(I_1, I_2)$, the SAM3 model is first applied with a fixed confidence threshold $c$ to segment the target categories of interest, yielding the corresponding sets of masks $(M_1, M_2)$. 
Subsequently, the Intersection over Union (IoU) between masks from different temporal instances is computed, and mask pairs are matched and filtered based on their IoU discrepancies. 
In this manner, regions exhibiting significant changes are identified and extracted, leading to the final change detection results. 

\newpage
\section{Other Features of \textit{UniVCD}}
\label{sec:other_features}

In this appendix, we present additional qualitative results to illustrate the behavior of \textit{UniVCD} on small-object changes and oblique photography.

\subsection{Small-Object Change Detection}

\begin{figure}[htbp]
    \centering
    \includegraphics[width=.95\columnwidth]{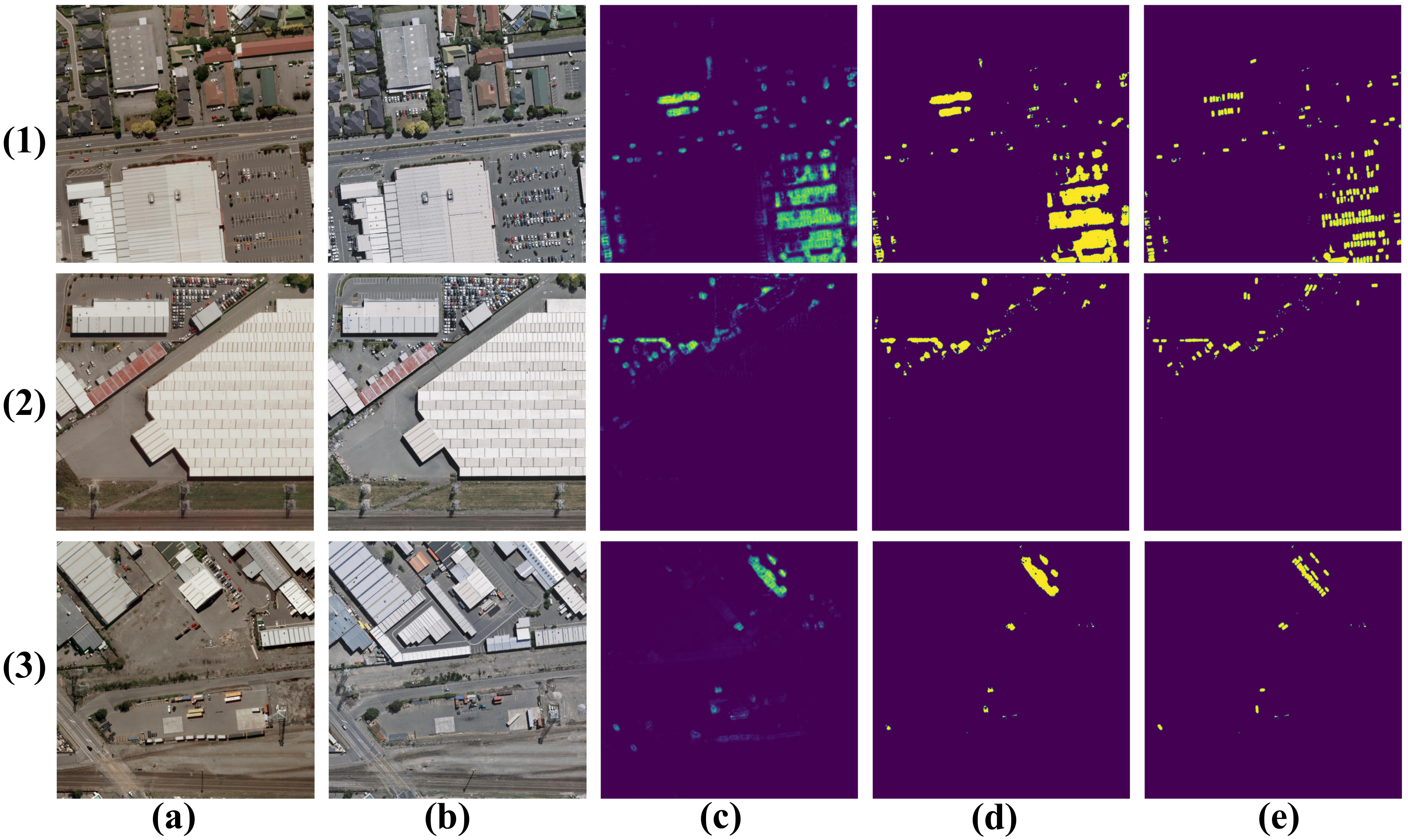}
    \caption{
    This figure illustrates the change detection performance of the proposed \textit{UniVCD} model for small objects (vehicles) on the WHU-CD\cite{WHUCD} dataset. Panels (a) and (b) show the pre- and post-change remote sensing images, while (c), (d), and (e) present the change-probability map, the binarized detection result, and the result after the post-processing step in Section~\ref{sam2 post}, respectively. As can be seen, \textit{UniVCD} can effectively detect changes of vehicles in the scene, demonstrating its potential for small-object change detection.
    }
    \label{fig:small_obj_cd}
\end{figure}

\subsection{Change Detection on Oblique photography}

\begin{figure}[htbp]
    \centering
    \includegraphics[width=.95\columnwidth]{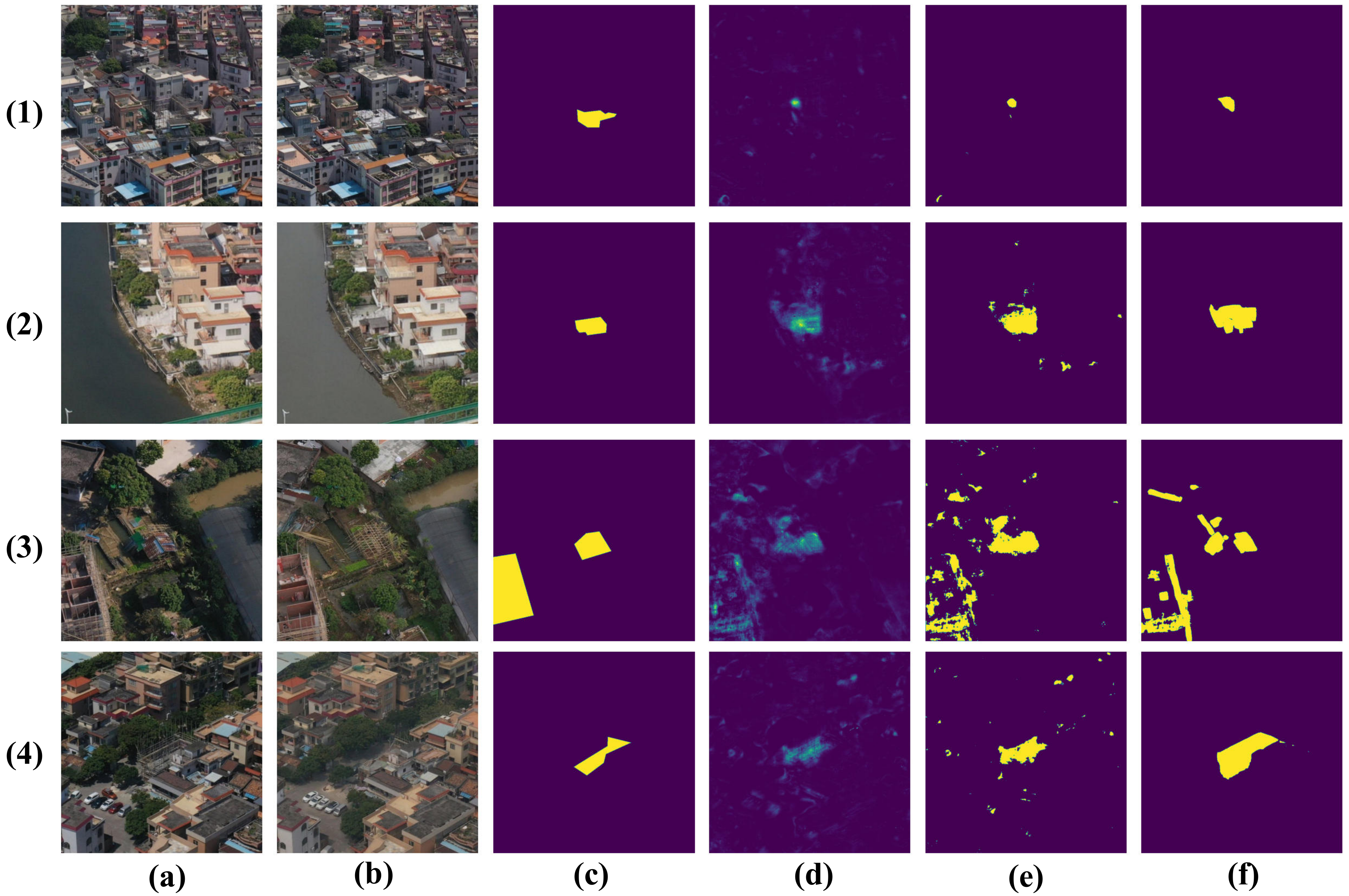}
    \caption{
    This figure illustrates the change detection performance of the proposed \textit{UniVCD} model on oblique photography. The image pair is taken from the UAV-BCD\cite{UAVBCD} dataset, which is designed for building change detection. Since many changes in this dataset involve building facades and thus do not fully satisfy the underlying assumptions of \textit{UniVCD} in Section~\ref{sec:post_limitations}, we report only qualitative results here. Panels (a) and (b) show the pre- and post-change UAV images, while (c), (d), (e), and (f) present the ground truth, the change-probability map, the binarized detection result, and the result after the post-processing step in Section~\ref{sam2 post}, respectively. As can be seen, \textit{UniVCD} can effectively detect building changes in oblique views, demonstrating its potential for this type of task.}
    \label{fig:uav_cd}
\end{figure}

\vfill
\end{document}